\documentclass[review]{iccv}

\usepackage{times}
\usepackage{epsfig}
\usepackage{graphicx}
\usepackage{amsmath}
\usepackage{amssymb}
\usepackage{caption}
\usepackage[ruled]{algorithm2e}
\usepackage[accsupp]{axessibility}
\usepackage[utf8]{inputenc}
\usepackage{color}
\usepackage{ifthen}
\usepackage{float}
\usepackage{alltt}
\usepackage{newlfont}
\usepackage{wrapfig}
\usepackage{booktabs}
\usepackage{multirow}
\usepackage{array}
\usepackage{appendix}
\usepackage{comment}
\usepackage{amsthm}
\usepackage{textcomp}
\usepackage{makecell}

\usepackage[pagebackref=true,breaklinks=true,colorlinks,bookmarks=false]{hyperref}
\iccvfinalcopy % *** Uncomment this line for the final submission

\def\iccvPaperID{2731} % *** Enter the ICCV Paper ID here

\def\confYear{ICCV 2021}

\definecolor{MyRed}{RGB}{0, 0, 0}

\graphicspath{
	{figs/raster/}
}

\begin{document}
	\title{Trash to Treasure: Harvesting OOD Data with Cross-Modal Matching \\ for
	Open-Set Semi-Supervised Learning}
%	\title{Joint Representation and Classifier Learning for \\ Open-Set Semi-supervised Classification}
	
	\author{Junkai Huang$^{1}$\footnotemark[1]\quad Chaowei Fang$^{2}$\footnotemark[1]\quad Weikai Chen$^{3}$\quad Zhenhua Chai$^{4}$\quad \\ Xiaolin Wei$^{4}$\quad Pengxu Wei$^{1}$\quad Liang Lin$^{1}$\quad Guanbin Li$^{1}$\footnotemark[2]\\
		$^{1}$Sun Yat-sen University\quad $^{2}$Xidian University\quad $^{3}$Tencent America\quad $^{4}$Meituan\\
% 		Institution1 address\\
% 		{\tt\small firstauthor@i1.org}
% 		\and
% 		Second Author\\
% 		Institution2\\
% 		First line of institution2 address\\
% 		{\tt\small secondauthor@i2.org}
	}

% \twocolumn[{%
\maketitle
%\thispagestyle{empty}
% \begin{center}
%     \centering
%     \includegraphics[width=\textwidth]{figures/teaser_1116.pdf}
%     \captionof{figure}{Our overall idea is ...}
%     \label{fig:teaser}
% \end{center}%
% }]
    \renewcommand{\thefootnote}{\fnsymbol{footnote}}
    \footnotetext[1]{Equal contribution.}
    \footnotetext[2]{Corresponding author.}

\begin{abstract}
   Open-set semi-supervised learning~(open-set SSL) investigates a challenging but practical scenario where out-of-distribution (OOD) samples are contained in the unlabeled data. While the mainstream technique seeks to completely filter out the OOD samples for semi-supervised learning (SSL), we propose a novel training mechanism that could effectively exploit the presence of OOD data for enhanced feature learning while avoiding its adverse impact on the SSL. We achieve this goal by first introducing a warm-up training that leverages all the unlabeled data, including both the in-distribution (ID) and OOD samples. Specifically, we perform a pretext task that enforces our feature extractor to obtain a high-level semantic understanding of the training images, leading to more discriminative features that can benefit the downstream tasks. Since the OOD samples are inevitably detrimental to SSL, we propose a novel cross-modal matching strategy to detect OOD samples.
   Instead of directly applying binary classification~\cite{yu2020multi}, we train the network to predict whether the data sample is matched to an assigned one-hot class label. The appeal of the proposed cross-modal matching over binary classification is the ability to generate a compatible feature space that aligns with the core classification task.  Extensive experiments show that our approach substantially lifts the performance on open-set SSL and outperforms the state-of-the-art by a large margin.
\end{abstract}

\section{Introduction}
\label{sec:intro}

\begin{figure}[t]
	\centering
	\includegraphics[width=0.95\linewidth]{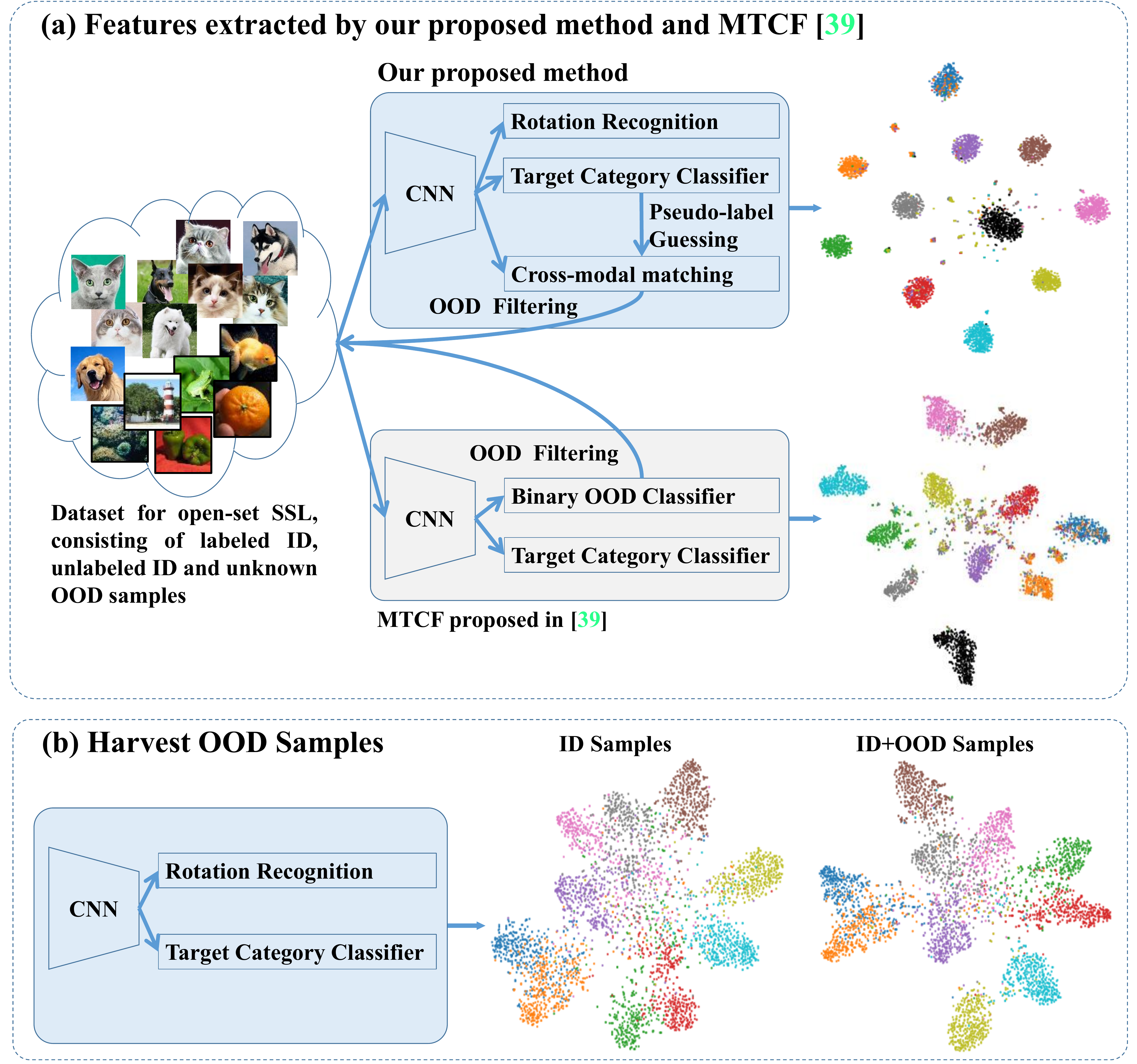}
	\caption{t-SNE~\cite{maaten2008visualizing} visualization of image features extracted from CIFAR100~\cite{krizhevsky2009learning}.
	Images of the same category are shown in the same color. (a) Features learned by our method are more compact and cleaner than those learned by MTCF~\cite{yu2020multi}.
	(b) By leveraging OOD samples in the proposed pretext training, we have achieved more discriminative feature space than using ID samples only.
	}
	\label{fig:teaser}
	\vspace{-3mm}
\end{figure}

\emph{``One man's trash is another man's treasure."}

\hspace{54mm}\emph{-- \small{Hector Urquhart}}

Semi-supervised learning~(SSL) provides an effective way of leveraging massive unlabeled data to improve the performance of deep neural network when only limited labeled samples are available.
Most existing SSL methods assume that labeled and unlabeled data share the same category space.
However, this assumption is difficult to satisfy since it still requires tedious efforts to confirm the purity of the unlabeled data.
Very recently, Yu \textit{et al.}~\cite{yu2020multi} proposed a more realistic setting called open-set semi-supervised learning~(open-set SSL).
Open-set SSL considers a more challenging but practical scenario where outliers, that do not belong to the categories of the labeled data, may exist in the unlabeled data.
Resolving the open-set SSL problem has crucial practical meanings as it can significantly reduce the workload of data preparation in the actual applications.

A straightforward approach to cope with the out-of-distribution (OOD) samples is to completely remove them from the SSL training, as the prior works~\cite{oliver2018realistic} have shown that including OOD unlabeled data can severely impact the performance of SSL.
While there exist a diverse collection of approaches for OOD sample detection, they typically require a large corpus of in-distribution (ID) data with class labels.
However, due to the scarcity of the labeled data in SSL, the existing OOD detection methods would fail to achieve satisfactory performance and are hence not suitable to be deployed in the open-set SSL.

In order to eliminate the influence of OOD samples,~\cite{yu2020multi} devises a multi-task curriculum framework (MTCF) with a binary OOD classification head that strives to filter out all the OOD samples.
The classification of in-distribution~(ID) samples and the detection of OOD samples are unified into a joint optimization framework, where unlabeled samples with lower OOD scores will be gradually added for semi-supervised training.
However, the proposed binary OOD classification task and the ID classification have conflicting goals in terms of feature learning.
Specifically, the training of OOD detection aims to cluster all ID samples (regardless of their categories) into one category (i.e., ID data) while the task of ID classification tends to enhance the category discrimination between ID samples.
Unifying the contradictory optimization goals into one framework that shares a backbone network could compromise the final performance and increase the difficulty of training.

In this paper, we present a novel training framework for open-set SSL that can effectively exploit the presence of OOD data for enhanced feature learning while avoiding its adverse impact to the SSL.
First, instead of completely discarding the OOD data, we introduce a warm-up training that makes full use of all the unlabeled data, including the OOD samples, to enhance the representation learning of our backbone network.
Unlike the conventional pre-training,  our warm-up training performs a pretext task that deviates from the target application.
In particular, we ask the network to predict the rotation of the rotationally augmented data in a self-supervised manner.
This enforces our backbone model to obtain high-level semantic understanding of the images and hence leads to more discriminative features that could benefit the downstream applications, e.g. the classification task.
Particularly, as shown in Figure~\ref{fig:teaser}(b), OOD samples, which are outliers in SSL algorithms, turn out to be treasures that can enhance feature learning when fully utilized in the proposed self-supervised pretext training.
The idea of leveraging self-supervised techniques for boosting the performance of semi-supervised learning has been shown effective in the previous work~\cite{Zhai_2019_ICCV}. However, it is only verified in the traditional SSL, where the unlabeled data share the same category space with the labeled ones.
We are the first to investigate this idea in the open-set setting and show that the self-supervised auxiliary task could be beneficial to open-set SSL with properly designed training strategy.

Second, we propose a more effective approach for detecting and filtering OOD samples based on a novel cross-modal matching mechanism.
First of all, each unlabeled sample is assigned the category with the highest predicted probability of the model as a pseudo label.
We then propose a cross-modal matching head to infer whether the embedding of the image and its pseudo-label are matched.
Once trained, OOD samples can be screened out due to its low confidence with all the ID categories.
Unlike the binary classification based OOD detection~\cite{yu2020multi}, the feature learning of cross-modal matching aligns well with that of the target ID classification task, as both strive to achieve better discrimination between the image features of different categories.
We show in Figure~\ref{fig:teaser}(a) that our method can obtain features with much more compact and purified clusters than that of \cite{yu2020multi}.
Furthermore, we can effectively detect ID samples with incorrect pseudo labels, also coded ``hard'' samples, via cross-modal matching.
This helps to further improve the performance of the trained model since hard samples could harm the model training especially at the early stage when the prediction accuracy of pseudo labels is relatively low.
We propose an adaptive training mechanism which gradually involves more hard samples as the model proceeds to achieve better performance.

Our proposed approach is a general training framework that can be easily implemented into existing SSL methods.
We show that our method greatly improves the state-of-the-art performance in extensive open-set semi-supervised image recognition benchmarks including CIFAR-10~\cite{krizhevsky2009learning}, Animals-10, CIFAR-100, and TinyImageNet~\cite{le2015tiny}.
We summarize our contributions as follows:
\vspace{-2mm}
\begin{itemize}
	\item A novel training pipeline for open-set SSL that leverages the presence of OOD samples for enhanced feature learning while avoiding their adverse impact.
	\vspace{-2mm}
	\item A specially tailored warm-up training method that uses self-supervised learning to boost the performance of open-set SSL.
	\vspace{-2mm}
	
	\item A novel OOD and hard sample detection algorithm based on cross-modal matching, which achieves compatible feature space with the target classification task.
	\vspace{-6mm}
	
	\item New state-of-the-art performance on open-set SSL over extensive benchmarks including CIFAR-10, CIFAR-100, TinyImageNet, and Animals-10.
\end{itemize}

\section{Related Works}
\label{sec:related_work}

\noindent \textbf{Semi-Supervised Learning.}
Semi-supervised learning~(SSL) generally refers to a series of methods that aim to use both labeled and unlabeled data from the similar distribution for model training. There are a vast number of classic works on SSL across various disciplines~\cite{kall2007semi,chen2014semi,kashima2009link}. Among them, semi-supervised image classification has been a long-term and extensive research topic. The most classic solution for the SSL problem is self-training, which iteratively enlarges the labeled set via guessing labels of unknown samples.
We refer interested readers to ~\cite{triguero2015self} for a detailed survey.
Other techniques including co-training~\cite{blum1998combining}, label propagation~\cite{Raghavan2007NearLT} and graphical model are also widely used in this task~\cite{Zhu2002LearningFL,wang2013dynamic,belkin2004semi,he2007graph}.

Benefiting from deep learning, breakthrough achievements have been achieved in SSL. Traditional semi-supervised techniques are re-implemented with deep CNN, such as self-labeling~\cite{triguero2015self}, multi-view training~\cite{qiao2018deep,dong2018tri}, label propagation~\cite{iscen2019label} and graph-based method~\cite{kipf2016semi}.  These methods focus on assigning pseudo hard/soft labels to unknown samples or clustering samples with similar semantics.
Considering the sample density is low nearby the decision boundary, a training sample is assumed to share the same label with a synthesized sample close to it.
Motivated by this intuition, consistency regularization can be exerted to spread labels of known samples and confident predictions of unknown samples.
In~\cite{laine2016temporal}, two models including $\pi$-model and temporal ensembling model are proposed to regularize the predictions on two different augmentations of a training sample.
Unlike the temporal ensembling model which averages the prediction in every iteration, the mean teacher~\cite{2017Mean} aggregates model weights through the exponential moving average.
MixMatch~\cite{berthelot} introduces mixup~\cite{zhang01} to explore inter-class relations in SSL.
\cite{xie2019unsupervised,Sohn2020FixMatch} employ two variants of augmentations and propagate the prediction of the weakly augmented image to the strongly augmented counterpart. Most of the mainstream SSL algorithms mentioned above are based on the assumption that the labeled and unlabeled samples share the category space. We break this assumption and focus on the new setting of open-set SSL.
\vspace{1mm}

\noindent \textbf{Self-supervised Learning.}
Self-supervised learning is an emerging technology which is widely considered to have the potential to initialize CNN with powerful representation capacity. It cleverly designs \textit{pretext} tasks, which can be formulated using only unlabeled data but requiring high-level semantic understanding. As a result, the intermediate layers of convolutional neural networks  trained for solving these \textit{pretext} tasks encode high-level semantic representation that can be universally applied to downstream tasks, e.g., image classification. The most commonly-used pretext tasks include transformation based regularization~\cite{Dosovitskiy2014discriminative}, patch based jigsaw puzzle~\cite{Noroozi2016jiasaw}, relative location inference~\cite{Doersch2015context}, and rotation recognition~\cite{gidaris2018unsupervised}, etc.  \cite{kolesnikov2019revisiting} revisits these self-supervised methods and provides comprehensive quantitative comparisons.
Recently, Zhai~\textit{et al.}~\cite{Zhai_2019_ICCV} proved that extra self-supervisions such as rotation recognition and transformation based regularization can benefit the classification performance in semi-supervised image classification.
However, it is under explored whether the performance gains obtained by self-supervision will be exhausted by the interference caused by OOD samples in the process of open-set SSL.
In this paper, rotation recognition is introduced as an auxiliary task to make full usage of all samples including OOD samples for feature representation enhancement.
\vspace{1mm}

\noindent \textbf{Open-set Semi-supervised Learning.}
Recently, researchers have gradually focused on settling the open-set SSL problem.
\cite{chen2020semi} proposes to solve the problem in which the categories of labeled and unlabeled samples are not completely matched. In~\cite{yu2020multi}, the concept of open-set SSL is put forward for the first time.
With the help of a binary OOD classifier, a curriculum framework is proposed to solve this problem. We argue that the OOD classification does not benefit to learn discriminative feature representations for the core category recognition task but actually have conflicting targets with the target category classification. Based on this concern, we propose a novel method to filter out OOD samples based on self-labeling and cross-modal matching between images and labels.
\textcolor{MyRed}{UASD~\cite{chen2020semi} temporally accumulates the network predictions for self-distillation, and uses a simple threshold on the largest prediction score to detect OOD samples. However, its detection of OOD data is highly sensitive to the performance of the final classifier. Our proposed cross-modal matching strategy removes the dependency on the outcome of the classifier by inferring whether the embedding of the input image is matched to an assigned pseudo-label. It also helps detect ID samples with incorrect pseudo labels (``hard” samples) that cannot be handled by UASD. DS3L~\cite{guo2020safe} introduces meta-learning to suppress the weight of OOD samples.} Instead of directly filtering out OOD samples, \cite{luo2021consistency} attempts to remove the distribution divergence between ID and OOD samples based on style transfer, and then explore the OOD samples during training via the unsupervised data augmentation~\cite{xie2019unsupervised}.
The distribution gap between ID and OOD samples is caused by category difference. Style transfer can only change the style of OOD images but not the semantic content. Hence, it remains difficult for style transfer technique to fully eliminate the feature discrepancy between the ID and OOD samples.

\begin{figure*}[t]
	\centering
	\includegraphics[width=0.8\linewidth]{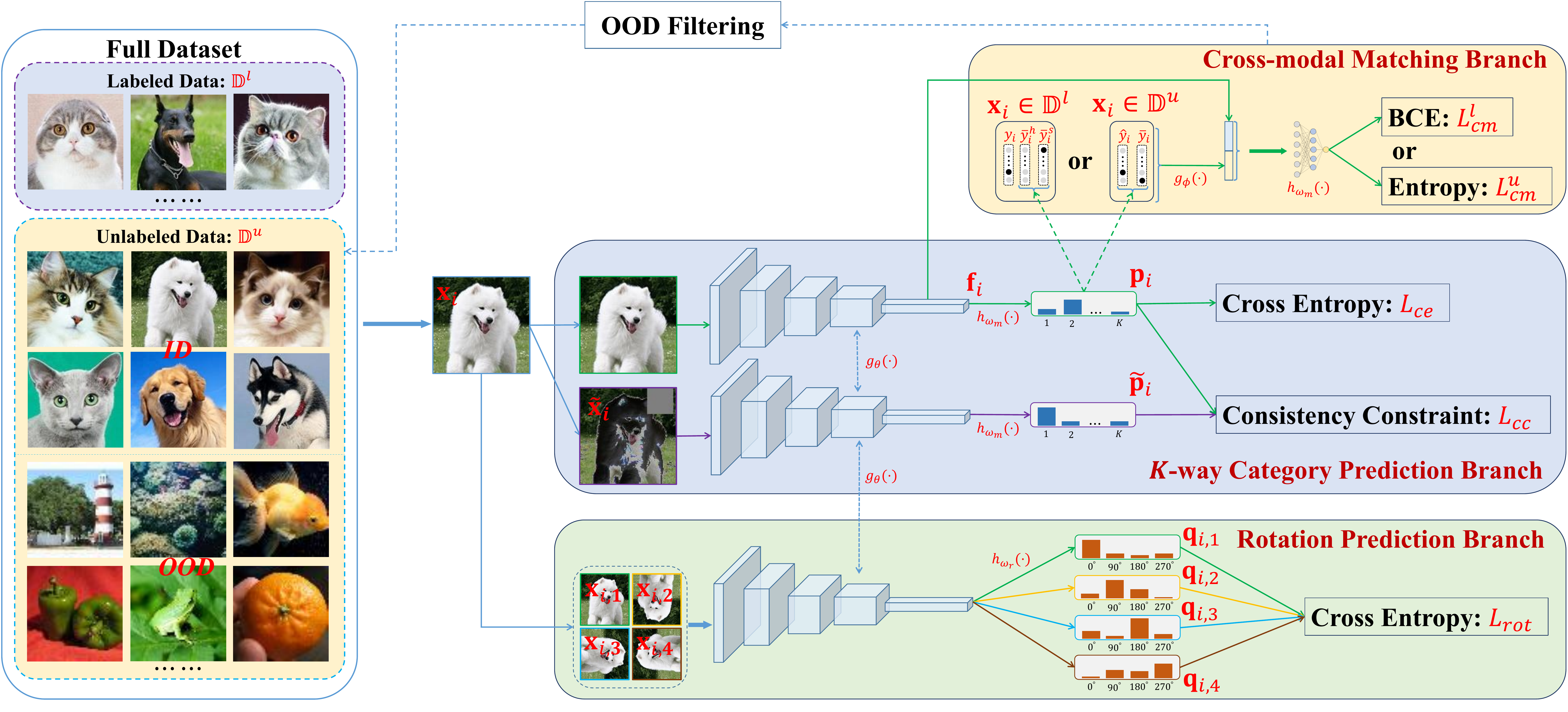}
	\caption{Overall architecture of our proposed method for open-set semi-supervised classification. It is composed of a multi-task framework, including a core category prediction branch, a rotation prediction branch for self-supervised feature learning, and a cross-modal matching branch for filtering out OOD samples within unlabeled data.}
	\label{fig:arch}
	\vspace{-3mm}
\end{figure*}

\section{Method}
\label{sec:method}

Similar to the SSL problem, the training dataset of the open-set SSL problem contains a small subset of labeled samples $\mathbb D^l=\left\{(\mathbf x_i^l, y_i^l)\right\}_{i=1}^{n=N}$ and a large subset of unlabeled samples $\mathbb D^u=\left\{\mathbf x_i^u\right\}_{i=1}^{M}$. Here, $\mathbf x_i^l$ or $\mathbf x_i^u$ represents a labeled or unlabeled image. $y_i^l$ denotes the ground-truth category of $\mathbf x_i^l$. Assuming that there are $K$ target categories, $y_i^l\in\{1,\cdots,K\}$. The `open-set' setting indicates that there exists OOD samples in the unlabeled training set. Namely, $\mathbf x_i^u$ may not belong to any one of the $K$ target categories.
We assume that a single training batch is consisting of $n$ labeled images and $m$ unlabeled images.
The overall framework of our method is illustrated in Figure \ref{fig:arch}.

\subsection{$K$-way Category Prediction} \label{sec:overall}

A convolutional neural network is chosen as the backbone for extracting feature representations from input images.
It is used to extract a 128-dimensional feature $\mathbf f_i$ from an input image $\mathbf x_i$, namely $\mathbf f_i = g_{\theta}(\mathbf x_i)$. $\theta$ represents the network parameters, and $g_{\theta}(\cdot)$ denotes the calculation function of the backbone model. To obtain the prediction scores for the $K$ target categories, a linear and a softmax layer is attached to the backbone, giving rise to a $K$-dimensional class probability vector $\mathbf p_i=h_{\omega_c}(\mathbf f_i)=h_{\omega_c}(g_{\theta}(\mathbf x_i))$. $\omega_c$ contains the weights and biases of the linear layer. During training, cross entropy is used to regularize the class probability vectors of labeled images,
\begin{equation}\label{eq:sup}
L_{ce}=-\frac{1}{n}\sum_{i=1}^n \ln(p_i^l[y_i^l]).
\end{equation}
Here, $p_i^l[k]$ indicates the $k$-th element of $\mathbf p_i^l$ which refers to the predicted probability vector of $\mathbf x_i^{l}$.
Inspired by~\cite{xie2019unsupervised}, the unsupervised consistency constraint is employed to pull close the distances between the predictions of every sample and its vicinal points. A strongly augmented counterpart $\tilde{\mathbf{x}}_i$ is synthesized for the training image $\mathbf{x}_i$.
Denote the category prediction of $\mathbf{x}_i$ and $\tilde{\mathbf{x}}_i$ be $\mathbf{p}_i$ and $\tilde{\mathbf{p}}_i$ respectively.
The KL divergence is adopted to calculate the distance between the two predictions, and the loss function of the consistency constraint is as below,
\begin{equation} \label{eq:losskl}
L_{cc}=\frac{1}{n+m}\sum_{i=1}^{n+m} \sum_{j=1}^K p_i[j]\ln(\frac{p_i[j]}{\tilde p_i[j]}).
\end{equation}

\subsection{Self-Supervised Representation Enhancement}
For the sake of enhancing the representation capacity of the backbone with all training samples including both ID and OOD samples, the rotation recognition is introduced as an auxiliary task.
In details, an extra 4-way rotation classification head consisting of 1 linear layer and 1 softmax function is attached to the backbone as shown in Figure \ref{fig:arch}.
We denote the calculation process of the rotation classification head as $h_{w_r}(\cdot)$, where $w_r$ represents related parameters.
For every training image, four counterparts are generated through rotating it by $0^\circ$, $90^\circ$, $180^\circ$ and $270^\circ$, respectively. %Then, they are fed through the backbone and the rotation prediction head.
We denote images generated via rotating $\mathbf x_i$ by $(j-1)*90^\circ$ as $\mathbf{x}_{i,j}$, and its rotation prediction as $\mathbf q_{i,j}=h_{w_r}(g_\theta(\mathbf{x}_{i,j}))$.
The following loss function is added during the training stage,
\begin{equation}\label{eq:rot}
L_{rot}=-\frac{1}{4(n+m)}\sum_{i=1}^{n+m} \sum_{j=1}^4 \ln(q_{i,j}[j]).
\end{equation}

This rotation prediction branch is critical to take advantage of large amounts of unlabed samples, especially OOD samples, to improve the representation learning.

\subsection{Cross-Modal Matching} \label{sec:cross}
In order to protect the learning of the $K$-way category recognition task from being distorted by OOD samples, a cross-modal matching branch is devised to purify unlabeled samples.
For an unlabeled sample $\mathbf{x}_i^u$, we assume the predicted probability vector be $\mathbf{p}_i^u$. The category having the most confident probability value is allocated as the pseudo label of $\mathbf{x}_i^u$, namely, $\hat y_i^u=\arg\max_j p_i^u[j]$.
The target of the cross-modal matching branch is trained to judge whether an ID image and a label from the set of target categories  are matched. It can be used to identify out OOD samples since they are not belonging to any of target categories, namely they are not matched to any target category.

Given a pair of an input sample $\mathbf{x}$ and a category label $y$, we first extract a feature vector $\mathbf f=g_\theta(\mathbf x)$ for $\mathbf{x}$. $y$ is transformed into a one hot vector which is subsequently transformed into a $128$-dimensional embedding vector via a linear layer, $\mathbf e=g_{\phi}(y)$. Afterwards, $\mathbf e$ and $\mathbf f$ are concatenated and fed into a multi-layer perceptron consisting of a hidden layer with $128$-dimensional output followed by the ReLU function and a linear layer attached with the Sigmoid function, giving rise to a matching score $s(\mathbf{x},y)=h_{\omega_m}(\mathbf f,\mathbf e)=h_{\omega_m}(g_\theta(\mathbf x),g_{\phi}(y))$. The matching score $s(\mathbf{x},y)$ measures whether $y$ is the correct category label of $\mathbf{x}$. When training the cross-modal matching head, positive samples can be easily collected from labeled images. Negative samples are synthesized through making pairs of images and categories which are not identical to the ground-truth label. For every training image, negative training samples are constructed in two manners: 1) Inspired by hard example mining~\cite{shrivastava2016training}, a so-called hardest label which is different from the ground-truth but having the largest prediction score is chosen; 2) The other relatively simple label is randomly selected from the category set excluding the ground-truth label and the mined hardest label. The following loss function is used for training the cross-modal matching head,
\begin{eqnarray} \label{eq:cross}
\nonumber L_{cm}^l =& -\frac{1}{n}\sum_{i=1}^n [\ln(s(\mathbf{x}_i^l,y_i^l))+\ln(1-s(\mathbf{x}_i^l, \bar{y}_i^{l,h})) \\
&+\ln(1-s(\mathbf{x}_i^l, \bar{y}_i^{l,s}))].
\end{eqnarray}
Here, $\bar y_i^{l,h}$ and $\bar y_i^{l,s}$ indicates the hardest and a relatively simple negative label, respectively,
\begin{eqnarray}
\label{eq:hard} \bar y_i^{l,h} &=& \arg \max_{y\neq y_i^l} p_i[y], \\
\label{eq:easy} \bar y_i^{l,s} &=& \textrm{rand}(\{y\in[1,K]\;|\;y\neq y_i^l;\;y\neq \bar y_i^{l,h}\}).
\end{eqnarray}

Considering the labeled samples are limited in open-set SSL, the unlabeled samples are employed to further strengthen the cross-modal matching head with entropy minimization~\cite{Grandvalet2005Semi},
\begin{eqnarray} \label{eq:crossu}
\nonumber L_{cm}^u &=& -\frac{1}{m}\sum_{i=1}^m [s(\mathbf{x}_i^u,\hat y_i^u)\ln(s(\mathbf{x}_i^u,\hat y_i^u)) + \\
\nonumber & & (1-s(\mathbf{x}_i^u, \hat{y}_i^{u}))\ln(1-s(\mathbf{x}_i^u, \hat{y}_i^{u})) +\\
\nonumber & & s(\mathbf{x}_i^u,\bar y_i^u)\ln(s(\mathbf{x}_i^u,\bar y_i^u))+ \\
& & (1-s(\mathbf{x}_i^u, \bar{y}_i^{u}))\ln(1-s(\mathbf{x}_i^u, \bar{y}_i^{u}))
],
\end{eqnarray}
where, $\bar y_i^{u} = \textrm{rand}(\{y\in[1,K]\;|\;y\neq \hat y_i^{u}\})$.

Apart from identifying OOD samples, the other utility of the cross-modal matching head is to preclude parts of misclassified ID samples. This is essential to reduce the instability of training the core classification task, especially when the classification performance of the model is still unfavorable in the early iterations.
The cross-modal matching branch is used to estimate the matching scores of all unlabeled samples and their pseudo labels inferred with the $K$-way category prediction branch.
The Otsu algorithm~\cite{Ostu1979A} is used to select the threshold for cleaning away samples with relatively low matching scores w.r.t their pseudo labels.

\subsection{Training Process}\label{sec:training}
The training process consists of two stages. In the first stage, we take a warm-up training stage for optimizing the complete architecture with loss function $L=L_{ce}+L_{cm}^l+L_{rot}$.
In the second stage, the cross-modal matching head is used to periodically clean unlabeled samples.
The consistency constraint (\ref{eq:losskl}) and entropy minimization (\ref{eq:crossu}) are added to train the $K$-way category prediction branch and the cross-modal matching branch, respectively.
The loss function of this stage is  $L=L_{ce}+L_{cm}^l+L_{cm}^u+L_{rot}+L_{cc}$. In such a manner, OOD samples are precluded after they are fully exploited for feature enhancement in the first stage.

\begin{table*}[t]
	\centering
	\fontsize{9}{10.8}\selectfont
	\setlength{\tabcolsep}{0.7mm}{
		\begin{tabular}{l|c|c|c|c|c|c|c|c|c|c|c|c}
			\toprule
			\multirow{2}{*}{Alg.} & \multicolumn{3}{c|}{TIN} & \multicolumn{3}{c|}{LSUN} & \multicolumn{3}{c|}{Gaussian Noise} & \multicolumn{3}{c}{Uniform Noise} \\ \cmidrule(l){2-4} \cmidrule(l){5-7} \cmidrule(l){8-10} \cmidrule(l){11-13}
			& 250  & 1000  & 4000  & 250  & 1000  & 4000  & 250  & 1000  & 4000  & 250  & 1000  & 4000  \\ \midrule
			MixM & $82.42_{0.70}$ & $88.03_{0.22}$ & $91.25_{0.13}$ & $76.32_{4.19}$ & $87.03_{0.41}$ & $91.18_{0.33}$  & $75.76_{3.49}$ & $85.71_{1.14}$ & $91.51_{0.35}$ & $72.90_{0.96}$ & $84.49_{1.06}$ & $90.47_{0.38}$ \\
			UDA & $88.83_{0.49}$ & $91.75_{0.12}$ & $93.63_{0.11}$ & $88.54_{1.10}$ & $91.12_{0.31}$ & $93.25_{0.12}$ & $88.93_{0.73}$ & $89.23_{0.41}$ & $92.35_{0.18}$ & $88.69_{0.93}$ & $89.74_{0.42}$ & $92.74_{0.35}$ \\
			%FixMatch~\cite{Sohn2020FixMatch} & $87.79 \pm 1.27$ & $91.54 \pm 1.76$ & $93.26 \pm 1.53$ & $88.13 \pm 0.56$ & $91.85 \pm 2.11$ & $93.58 \pm 1.39$ \\
			UASD & $83.53$ & - & - & $80.87$ & - & - & - &- &- &- &- &- \\
			DS3L & - & $70.10_{0.47}$ & $82.58_{0.14}$ & - & $69.74_{0.08}$ & $82.89_{0.69}$ & - & $62.86_{0.67}$ & $80.44_{0.01}$ & - & $62.89_{1.65}$ & $80.59_{0.03}$ \\
			MTCF & $86.44_{0.64}$ & $89.85_{0.11}$ & $93.03_{0.05}$ & $86.65_{0.41}$ & $90.19_{0.47}$ & $92.91_{0.03}$ & $87.34_{0.13}$ & $89.80_{0.26}$ & $92.53_{0.08}$ & $85.54_{0.11}$ & $89.87_{0.08}$ & $92.83_{0.04}$ \\
			OTCT & - & $91.10_{0.65}$ &$93.84_{0.10}$ & - & $91.30_{0.36}$ & $94.27_{0.21}$ & - & $92.33_{0.59}$ &$\color{red}{94.52}_{0.07}$ & - & $91.82_{0.04}$ & $94.50_{0.13}$ \\
			\midrule
			
			Ours & $\color{red}{91.52}_{0.11}$ & $\color{red}{93.26}_{0.14}$ & $\color{red}{94.71}_{0.06}$ & $\color{red}{91.13}_{0.21}$ & $\color{red}{94.43}_{0.10}$ & $\color{red}{94.97}_{0.10}$ & $\color{red}{90.81}_{0.12}$ & $\color{red}{93.63}_{0.06}$ & $94.38_{0.12}$ & $\color{red}{89.95}_{0.18}$ & $\color{red}{94.12}_{0.14}$ & $\color{red}{94.83}_{0.15}$ \\ \bottomrule
		\end{tabular}
	}
	\caption{\textcolor{MyRed}{Comparisons with the conventional SSL and open-set SSL algorithms, including MixM (short for MixMatch)~\cite{berthelot}, UDA~\cite{xie2019unsupervised}, UASD~\cite{chen2020semi} DS3L~\cite{guo2020safe}, MTCF~\cite{yu2020multi}, and OTCT~\cite{luo2021consistency},  on variants of CIFAR-10 which are respectively corrupted with two real-world OOD datasets (TIN and LSUN) and two synthetic OOD datasets (Gaussian Noise and Uniform Noise).
	Accuracy (\%) is used for evaluating algorithms.
	The subscript of the accuracy value indicates its standard deviation. 250, 1000, and 4000 labeled images are used for training respectively.
	We use the reported results of \cite{yu2020multi} for UASD and that of \cite{luo2021consistency} for DS3L and OTCT.
	}
	}
	\label{tab:cifar10real}
	\vspace{-3mm}
\end{table*}

\section{Experiments}
\label{sec:experiments}
\subsection{Datasets}
\label{sec:dataset}

The following four public datasets are used to validate the performance of open-set SSL algorithms.% described as follows.

\noindent \textbf{CIFAR-10}~\cite{krizhevsky2009learning} consists of 60,000 images of size $32\times32$ which belong to 10 categories. Following the original split, 10,000 images are used for testing.
The same splits in~\cite{yu2020multi} are adopted for training and validating. The number of labeled training images varies in \{250, 1,000, 4,000\}.

\noindent \textbf{Animals-10} which is obtained from Kaggle, contains 26,179 images of 10 animal categories.
500/1,000, 1,000 and 2,000  images are selected as labeled training, validating,  and testing samples respectively.
The remaining images are used as unlabeled samples.

\noindent \textbf{CIFAR-100} has 100 classes and each class contains 600 images. We choose 25,000 images of the first 50 classes as in-distribution samples, forming the CIFAR-ID-50 dataset. They are split into 22,500 samples for training and 2,500 samples for validating. The number of labeled training images is 2,000 or 2,500.

\noindent \textbf{TinyImageNet} (TIN)~\cite{le2015tiny} is composed of 120,000 images belonging to 200 classes. Similar to CIFAR-100, 27,500 images of the first 50 categories are regarded as in-distribution samples, which are separated into  22,500 samples for training, 2,500 samples for validating, and 2,500 samples for testing. We name the above subset of TIN as TIN-ID-50. In experiments below, 2,000/2,500 training images are selected as labeled samples.

The following inter-dataset and intra-dataset OOD settings are used in our experiments.

\noindent \textbf{Inter-Dataset OOD Setting} For the CIFAR-10 dataset, we follow~\cite{yu2020multi} to synthesize OOD samples. 10,000 images are sampled from each of the TIN dataset, the Large-scale Scene Understanding (LSUN) dataset~\cite{yu2015lsun}, Gaussian noise dataset, and uniform noise dataset, forming into 4 OOD settings. For the Animals-10 and CIFAR-ID-50 datasets, 10,000 images from TIN are used as OOD samples.

\noindent \textbf{Intra-Dataset OOD Setting} For CIFAR-ID-50, we select 100 images from each of the other 50 classes of CIFAR-100 as OOD images. For TIN-ID-50, 50 images from each of the other 150 classes of TIN are chosen as OOD images.

\subsection{Implementation details}
\label{sec:imp_details}

Existing methods including MixMatch~\cite{berthelot}, UDA~\cite{xie2019unsupervised} %, FixMatch~\cite{Sohn2020FixMatch}
\textcolor{MyRed}{FixMatch~\cite{Sohn2020FixMatch}, UASD~\cite{chen2020semi}, DS3L~\cite{guo2020safe}, OTCT~\cite{luo2021consistency}}
and MTCF~\cite{yu2020multi}, are used for comparison.
For UDA, \textcolor{MyRed}{FixMatch} and our method, SGD is used to optimize network weights. The learning rate is initially set to 0.03 and adjusted via the cosine decay strategy~\cite{xie2019unsupervised,Sohn2020FixMatch}. The momentum is set to 0.9. %The cosine decay strategy is used to adjust the learning rate,  following .
In each training batch, $n=64$, and $m=320$. For our method, the first stage costs 50,000 iterations, and the second stage takes 200,000 iterations.
Without specification, the cycle length of using the crossmodal matching head to clean unlabeled data is $2\times10^4$.
For UDA and FixMatch, models are trained with 250,000 iterations for a fair comparison.
When training MixMatch and MTCF, we follow the original settings of~\cite{yu2020multi} in which models are trained with 1,024 epochs, and each epoch contains 1,024 iterations.
During the training stage, network weights are saved every 1,000 iterations. The averaged classification accuracy of the last 20 copies is used to evaluate the performance of all methods.  For all experiments, we use the Wide-ResNet28-2~\cite{zagoruyko2016wide} as the backbone model.

\begin{table}[t]
\centering
\fontsize{9}{10.8}\selectfont
\setlength{\tabcolsep}{2.8mm}{
    \begin{tabular}{l|c|c|c|c}
        \toprule
        \multirow{2}{*}{Method} & \multicolumn{2}{c|}{Animals-10} & \multicolumn{2}{c}{CIFAR-ID-50} \\ \cmidrule(l){2-3}\cmidrule(l){4-5}
          & 500  & 1000  & 2000  & 2500  \\ \midrule
        MixMatch~\cite{berthelot} & $78.35$ & $83.15$ & $62.10$ & $64.78$ \\
        UDA~\cite{xie2019unsupervised} & $83.30$ & $84.74$ & $64.34$ & $66.65$ \\
        MTCF~\cite{yu2020multi} & $73.50$ & $75.83$ & $63.22$ & $65.10$ \\
        MTCF+UDA & $79.85$ & $85.60$ & $65.20$ & $67.30$ \\
        FixMatch~\cite{Sohn2020FixMatch} & $89.06$ & $91.00$ & $68.98$ & $72.92$ \\ \hline
        Ours+UDA & $87.86$ & $89.70$ & $71.58$ & $73.19$ \\
        Ours+FixMatch & $\mathbf{89.43}$ & $\mathbf{91.50}$ & $\mathbf{72.06}$ & $\mathbf{73.80}$ \\
        \bottomrule
    \end{tabular}
}
\caption{Accuracy (\%) for Animals-10 and CIFAR-ID-50. Images of TIN are used as OOD samples. On Animals-10 dataset, 500 or 1000 labeled images are used for training. On CIFAR-ID-50 dataset, 2000 or 2500 labeled images are used for training. }
\label{tab:inter}
\vspace{-3mm}
\end{table}

\begin{table}[t]
\centering
\fontsize{9}{10.8}\selectfont
\setlength{\tabcolsep}{2.8mm}{
    \begin{tabular}{l|c|c|c|c}
        \toprule
        \multirow{2}{*}{Method} & \multicolumn{2}{c|}{CIFAR-ID-50} & \multicolumn{2}{c}{TIN-ID-50} \\ \cmidrule(l){2-3}\cmidrule(l){4-5}
          & 2000  & 2500  & 2000  & 2500  \\ \midrule
        MixMatch~\cite{berthelot} & $60.20$ & $66.17$ & $48.12$ & $50.52$ \\
        % \midrule
        % \cmidrule{2-5}
        UDA~\cite{xie2019unsupervised} & $66.02$ & $67.82$ & $54.03$ & $55.39$ \\

        MTCF~\cite{yu2020multi} & $63.48$ & $65.38$ & $49.64$ & $52.08$ \\
        MTCF+UDA & $60.06$ & $63.08$ & $47.32$ & $51.20$ \\
        % \midrule
        FixMatch~\cite{Sohn2020FixMatch} & $68.06$ & $71.01$ & $56.82$ & $60.33$ \\ \hline
        Ours+UDA & $67.36$ & $69.14$ & $54.87$ & $57.08$ \\
        Ours+FixMatch & $\mathbf{68.65}$ & $\mathbf{73.14}$ & $\mathbf{57.48}$ & $\mathbf{62.64}$ \\
        \bottomrule
    \end{tabular}
}
\caption{Accuracy (\%) for CIFAR-ID-50 and TIN-ID-50 under the intra-dataset OOD setting.  On both datasets, 2000 or 2500 labeled images are used for training.}
\label{tab:intra}
\vspace{-3mm}
\end{table}

\begin{figure}[t]
	\centering
	\includegraphics[width=0.8\linewidth]{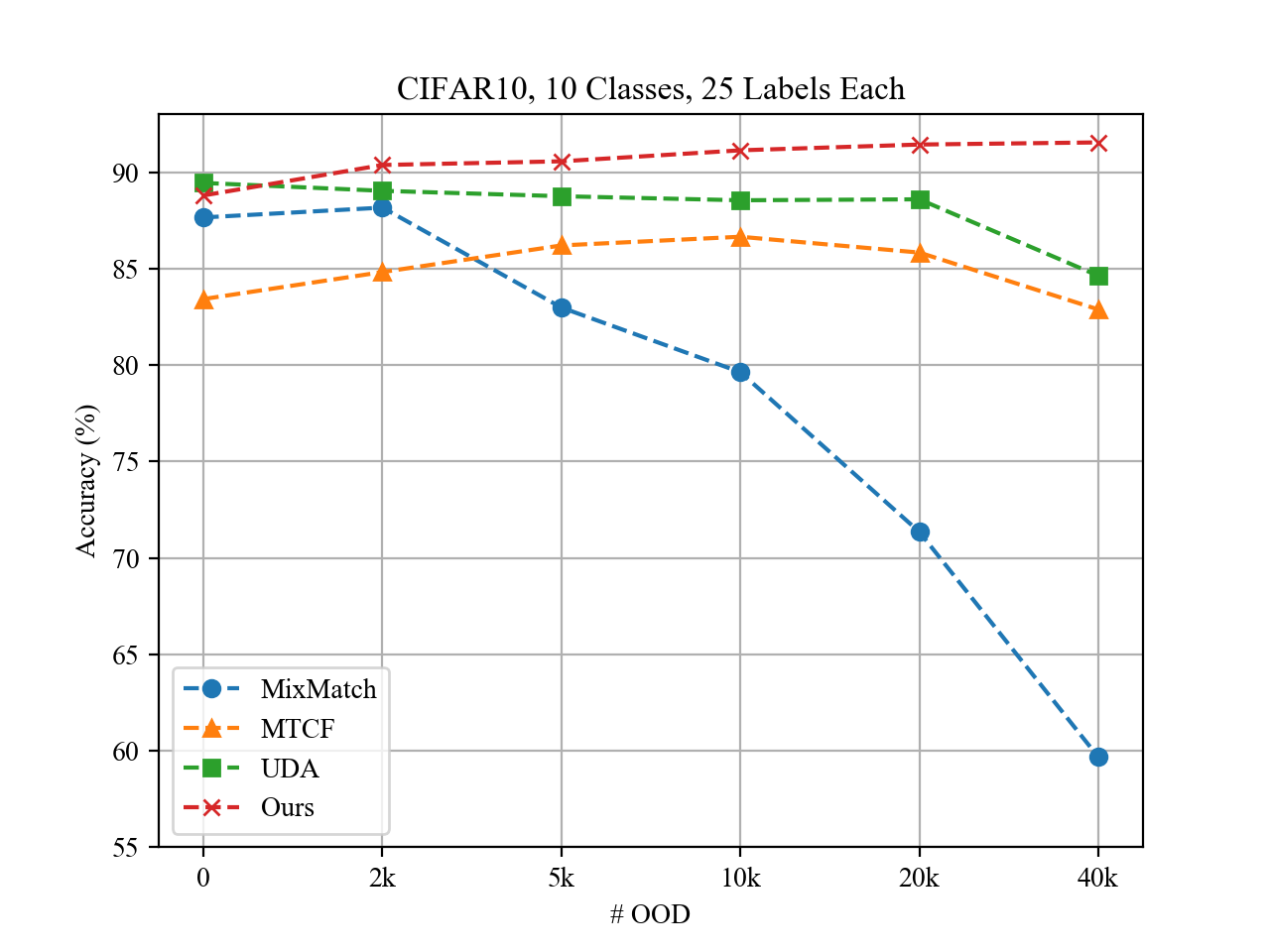}
	\caption{
	    Impact of different number of OOD samples on baselines and our approach.
	}
	\label{fig:num_of_ood}
	\vspace{-3mm}
\end{figure}

\subsection{Experimental Results}
\label{sec:exp_results}

\subsubsection{Comparison with Other Methods}
\label{sec:comparison}

\noindent \textbf{Inter-Dataset OOD Setting}
As introduced in Section \ref{sec:dataset}, three datasets including CIFAR-10 (Table \ref{tab:cifar10real}), Animals-10 (Table \ref{tab:inter}), and CIFAR-ID-50 (Table \ref{tab:inter}) are used to validate the classification performance under the inter-dataset OOD setting. %\textcolor{MyRed}{For UASD, we borrow its results on CIFAR-10 from~\cite{yu2020multi}. For DS3L and OTCT, we directly report their results from~\cite{luo2021consistency}.}
For all inter-dataset OOD settings of CIFAR-10, our method surpasses all compared methods by significant margins. As shown in Table \ref{tab:cifar10real}, when 250 labeled images are provided and images of TIN are used as OOD samples, our method achieves an average accuracy of 91.52\% which is 5.08\% and 3.69\% higher than MTCF and UDA, respectively.
Compared with two synthetic noise OOD datasets, the rich semantic information of real-world datasets is beneficial to our proposed method. For instance, the accuracy of our method achieves 91.13\% on the LSUN OOD setting, which is 1.28\% higher than the accuracy on the uniform noise OOD setting.

Table~\ref{tab:inter} reports the results of Animals-10 and CIFAR-ID-50 under the inter-dataset OOD setting in which images of TIN are added as OOD samples. \textcolor{MyRed}{Our method is capable of improving existing SSL methods, e.g. UDA and FixMatch, and performs much better than MTCF.}

We further study the impact of the number of OOD samples on SSL and open-set SSL algorithms. In this experiment, we use CIFAR-10 as the ID dataset and LSUN as the OOD dataset. The number of OOD samples varies from 0 to $4\times10^4$. %$\{0\textrm{,}\;2\times10^3\textrm{,}\;5\times10^3\textrm{,}\;1\times10^4\textrm{,}\;2\times10^4\textrm{,}\;4\times10^4\}$
25 labeled samples are provided for each class
The results are presented in Figure \ref{fig:num_of_ood}.
Our method achieves better performance with even more OOD samples, and consistently outperforms all the other methods in all settings.

\vspace{1mm}
\noindent \textbf{Intra-Dataset OOD Setting} The CIFAR-ID-50 and TIN-ID-50 datasets are adopted in this experiment. Images of the same dataset but having categories different from those ID categories are regarded as OOD samples.
The experimental results are shown in Table~\ref{tab:intra}. Our method consistently performs better than other methods.

We try to modify MTCF~\cite{yu2020multi} via using UDA~\cite{xie2019unsupervised} for semi-supervised learning, which forms a new open-set SSL algorithm denoted by `UDA+MTCF'. In MTCF, the learning of the target category classifier  is in conflict with the learning of the binary OOD classifier, thus the features learned by MTCF is not discriminative enough.
Meanwhile, UDA depends on the label propagation from the weakly augmented image to the strongly augmented counterpart, which is sensitive to a confused feature space.
The incorporation of MTCF only brings marginal improvement (e.g. CIFAR-ID-50 in Table~\ref{tab:inter}) or even causes severe performance degradation (e.g. both datasets in Table~\ref{tab:intra}) to UDA.

\begin{table}[t]
	\centering
	\fontsize{9}{10.8}\selectfont
	\setlength{\tabcolsep}{1.5mm}{
		\begin{tabular}{l|l|c|c|c|c}
			\toprule
			ID & OOD & ODIN & SUF & MTCF & Ours \\
			\midrule
			CIFAR-10    & LSUN     & $98.47$  & $99.03$ & $99.82$ & $\mathbf{99.98}$ \\
			Animals-10  & TIN     & $76.35$ & $90.01$ & $92.59$ & $\mathbf{93.51}$ \\
			CIFAR-ID-50 & TIN      & $88.82$ & $97.98$ & $98.17$ & $\mathbf{99.85}$ \\
			CIFAR-ID-50 & CIFAR-50 & $69.47$ & $72.32$ & $69.75$ & $\mathbf{74.13}$ \\
			TIN-ID-50   & TIN-150  & $59.83$ & $65.59$ & $63.92$ & $\mathbf{65.67}$ \\
			\bottomrule
		\end{tabular}
	}
	\caption{The comparison of our method against ODIN~\cite{liang2017enhancing},  SUF~\cite{lee2018simple}, and MTCF~\cite{yu2020multi} on the task of OOD detection. The evaluation metric is AUROC(\%). ODIN and SUF are implemented based on the classifier model learnt by UDA.
	For CIFAR-10 and Animals-10, 250 labels and 500 labels are used during training respectively. For other datasets, 2500 labels are used during training. }
	\label{tab:ood_detection}
	\vspace{-3mm}
\end{table}
\vspace{1mm}
\noindent \textbf{OOD Detection Performance}  In Table \ref{tab:ood_detection}, we compare our method against ODIN~\cite{liang2017enhancing}, Mahalanobis~\cite{lee2018simple} and MTCF~\cite{yu2020multi} under extensive combinations of ID and OOD datasets, to validate the efficacy of the cross-modal matching branch in detecting OOD images.
In our method, the matching score of an image and its pseudo label is regarded as the probability value of the image belonging to ID samples. The area under the receiver operating characteristic (AUROC) is used to measure the performance of OOD detection algorithms.
Our method outperforms ODIN, Mahalanobis and MTCF under all settings.

\begin{figure}[t]
	\centering
	\begin{minipage}{0.32\linewidth}
		\centering
		\centerline{\includegraphics[width=\linewidth,clip,trim=10 12 25 5]{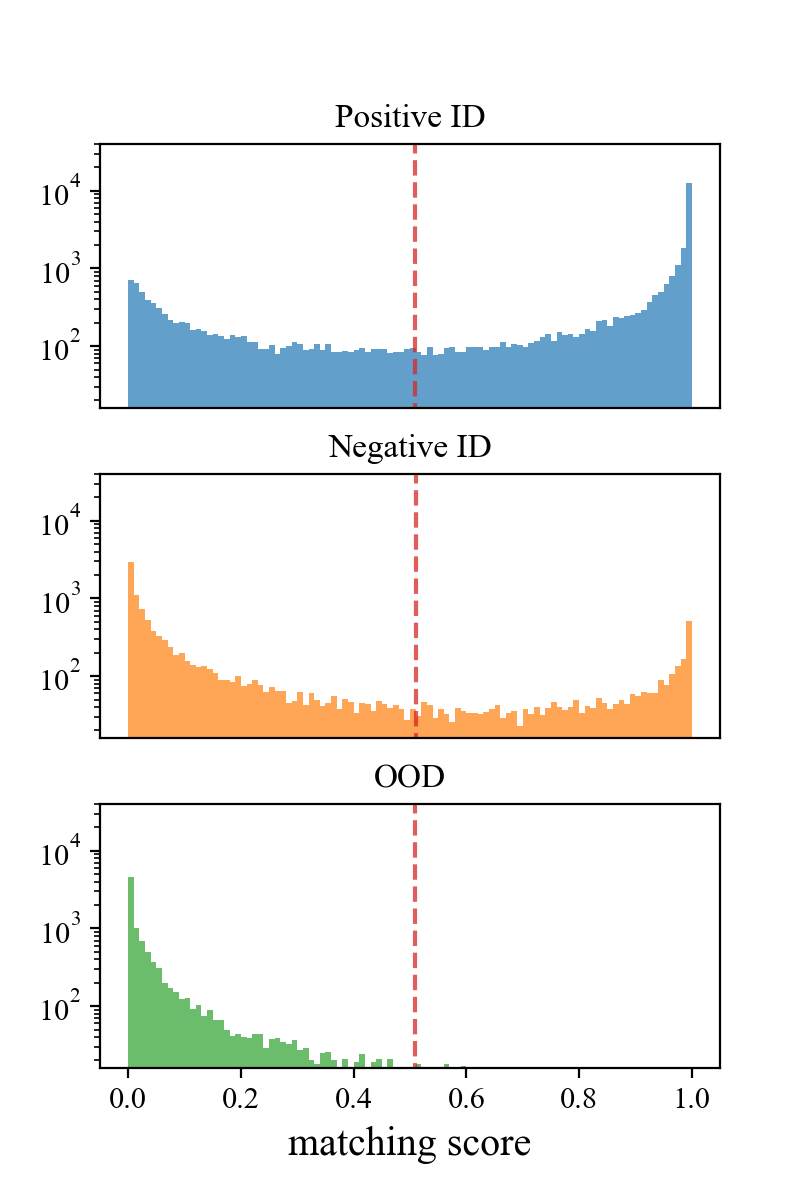}}
		\centerline{\footnotesize{1000-th iteration}}\medskip
	\end{minipage}
	\begin{minipage}{0.32\linewidth}
		\centering
		\centerline{\includegraphics[width=\linewidth,clip,trim=10 12 25 5]{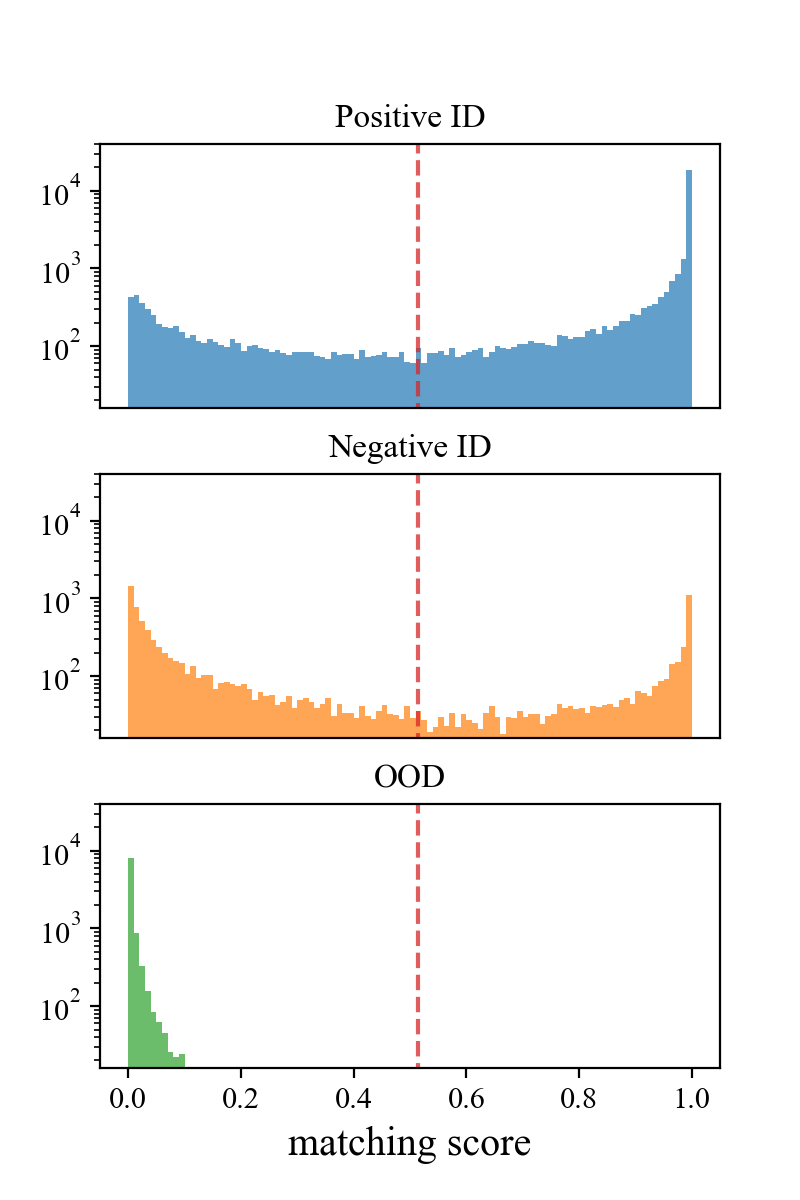}}
		\centerline{\footnotesize{19000-th iteration}}\medskip
	\end{minipage}
	\begin{minipage}{0.32\linewidth}
		\centering
		\centerline{\includegraphics[width=\linewidth,clip,trim=10 12 25 5]{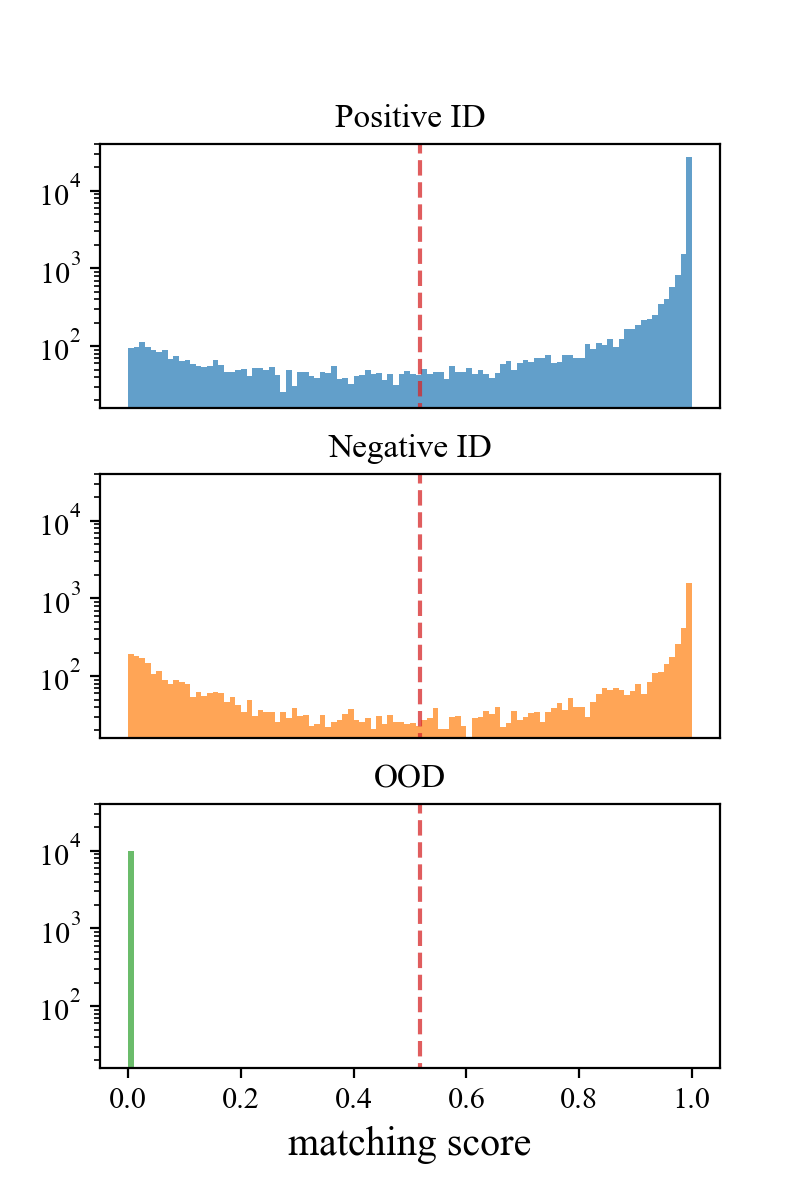}}
		\centerline{\footnotesize{28000-th iteration}}\medskip
	\end{minipage}
	\caption{Visualization of the matching scores in different iterations. The number of samples are showcased in logarithmic coordinates (vertical axis) with base of 10.  `Positive ID'/`Negative ID' means ID samples having correct/incorrect classification results. The threshold value obtained via OTSU is indicated by the dashed vertical line.}
	\label{fig:vismatching}
	\vspace{-3mm}
\end{figure}

\subsubsection{Ablation Study}
\label{sec:ablation}

\begin{table}[t]
	\centering
	\fontsize{9}{10.8}\selectfont
	\setlength{\tabcolsep}{1.5mm}{
		\begin{tabular}{c|l|cc}
			\toprule
			M-\# & Settings & CIFAR-10 & CIFAR-ID-50 \\ \midrule
			1 & Supervised & $39.66$ & $53.18$ \\
			2 & Supervised+SS & $75.69$ & $59.82$ \\
			3 & UDA  & $88.83$ & $64.78$ \\
			4 & UDA+SS  & $90.69$ & $70.10$ \\ \cmidrule{2-4}
			5 & UDA+SS+CT & 90.81 & 71.16 \\
			6 & UDA+SS+CMF-w/o-$L_{cm}^u$  & $91.09$ & $72.38$ \\
			7 & UDA+SS+CMF-w/o-WU  & $89.45$ & $71.81$ \\
			8 & UDA+SS+CMF-full  & $\mathbf{91.52}$ & $\mathbf{73.19}$ \\
			\bottomrule
		\end{tabular}
	}
	\caption{Ablation Study on CIFAR-10 and CIFAR-ID-50.  TIN is chosen as the OOD dataset. 250 and 2500 labeled images are for CIFAR-10 and CIFAR-ID-50, respectively. Abbreviations: unsupervised data augmentation (UDA)~\cite{xie2019unsupervised}, crossmodal matching based filtering (CMF),  self-supervision based on rotation recognition (SS), warming-up (WU), confidence-based thresholding (CT).
	}
	\label{tab:ablation2}
	\vspace{-3mm}
\end{table}
We carry out an extensive ablation study on the CIFAR-10 and CIFAR-ID-50 to tease apart the experimental factors that are most important to the success of our method, as shown in Table~\ref{tab:ablation2}.

\vspace{1mm}
\noindent \textbf{Self-supervised Learning in Open-set SSL} \textcolor{MyRed}{M-2 and M-4 in Table~\ref{tab:ablation2} represent the variant of the simple supervised learning and UDA respectively, which employs rotation recognition as the auxiliary task.
The accuracy of M-2 and M-4 on CIFAR-ID-50 is 5.32\% and 6.64\% higher than that of M-1 and M-3 respectively.} This indicates that the feature enhancement brought by the rotation recognition task can significantly improve the classification accuracy. This is the first time to prove that self-supervised representation learning based on rotation recognition is effective in open-set SSL.
\textcolor{MyRed}{We have tried other self-supervised learning methods, like SimCLR~\cite{chen2020simple} and MoCo~\cite{he2020momentum}, and empirically find that they perform much worse than the rotation prediction approach. Both SimCLR and MoCo strive to distinguish the features of different samples while our task requires samples of the same class to stay closer in the feature space. The conflicting goals make them unsuitable for our task.}

\vspace{1mm}
\noindent \textbf{Different Variants of OOD Sample Filtering}
The full version of our method (M-8 in Table \ref{tab:ablation2}) outperforms M-4 without cross-modal matching head by 3.09\% and 0.83\% on CIFAR-ID-50 and CIFAR-10 respectively. This validates that our proposed OOD sample filtering algorithm based on cross-modal matching is complementary to self-supervised learning, and it further benefits the classification performance of in-distribution categories.
On the other hand, merely using the supervised loss function (\ref{eq:cross}) to train the cross-modal matching branch (M-6) leads to performance degradation to a  certain extent. The accuracy on CIFAR-ID-50 is decreased by 0.81\%, compared to the results of M-8 in which both function (\ref{eq:cross}) and (\ref{eq:crossu}) are adopted for training. \textcolor{MyRed}{We also attempt to replace the cross-modal matching based filtering with the confidence-based thresholding via precluding unlabeled samples with maximum prediction scores lower than the OTSU threshold .
The resulted method (M-5) achieves the accuracy of 90.81\% and 71.16\% on CIFAR10 and CIFAR-ID-50 respectively, which is much inferior to that of our method. This indicates the superiority of the proposed cross-modal matching based filtering to the confidence-based thresholding, since the former can simultaneously implement the OOD sample filtering and promote the discrimination between features of different categories. }

\vspace{1mm}
\noindent \textbf{Efficacy of Warming Up Stage}
The warm-up stage plays a crucial role in fully leveraging all ID and OOD samples for representation learning, providing a decent initialization for the $K$-way category prediction and the cross-modal matching tasks.
Without the warm-up stage (M-7 in Table \ref{tab:ablation2}), the classification accuracy is reduced by 2.07\% and 1.38\% on CIFAR-10 and CIFAR-ID-50, respectively.

\subsubsection{Justification of Cross-modal Matching}
The performance of the cross-modal matching in identifying OOD samples and misclassified ID samples at different training iterations is visualized in Figure~\ref{fig:vismatching}.
The cross-modal matching head can effectively separate correctly classified samples from OOD samples, throughout the training process.
In early iterations, the head is capable of precluding most of the misclassified samples. This helps prevent these misclassified samples from misleading the optimization process. As the iteration progresses, the core classifier gradually becomes better and the number of correctly classified samples continues to increase.

\section{Conclusions}

In this paper, we propose a novel framework for open-set SSL which can harvest the OOD samples for enhanced feature learning while avoiding its adverse impact on the SSL training.
Our key insight is that OOD samples, if exploited in a pretext task of rotation recognition, can be ``treasures" for learning more discriminative features that can benefit the final classification task.
After the value of OOD samples in representation learning has been explored, the cross-modal matching branch is further utilized to filter out OOD samples without causing much interference to the core category prediction task.
Our proposed method can be easily integrated into existing SSL algorithms, and achieves state-of-the-art in open-set SSL on extensive public benchmarks.

\vspace{1mm}
\noindent \textbf{Acknowledgments}
This work was supported in part by the Guangdong Basic and Applied Basic Research Foundation under Grant No.2020B1515020048, in part by the National Natural Science Foundation of China under Grant No.61976250, No.U1811463 and No.62003256, and in part by the Guangzhou Science and technology project under Grant No.202102020633.
	%\fi
	
	{\small

	}


\begin{thebibliography}{10}\itemsep=-1pt

\bibitem{belkin2004semi}
Mikhail Belkin and Partha Niyogi.
\newblock Semi-supervised learning on riemannian manifolds.
\newblock {\em Machine learning}, 56(1):209--239, 2004.
\bibitem{berthelot}
David Berthelot, Nicholas Carlini, Ian Goodfellow, Nicolas Papernot, Avital
  Oliver, and Colin~A Raffel.
\newblock Mixmatch: A holistic approach to semi-supervised learning.
\newblock In {\em Advances in Neural Information Processing Systems}, pages
  5049--5059, 2019.

\bibitem{blum1998combining}
Avrim Blum and Tom Mitchell.
\newblock Combining labeled and unlabeled data with co-training.
\newblock In {\em Proceedings of the eleventh annual conference on
  Computational learning theory}, pages 92--100, 1998.

\bibitem{chen2020simple}
Ting Chen, Simon Kornblith, Mohammad Norouzi, and Geoffrey Hinton.
\newblock A simple framework for contrastive learning of visual
  representations.
\newblock In {\em International conference on machine learning}, pages
  1597--1607. PMLR, 2020.

\bibitem{chen2014semi}
Xing Chen and Gui-Ying Yan.
\newblock Semi-supervised learning for potential human microrna-disease
  associations inference.
\newblock {\em Scientific reports}, 4:5501, 2014.

\bibitem{chen2020semi}
Yanbei Chen, Xiatian Zhu, Wei Li, and Shaogang Gong.
\newblock Semi-supervised learning under class distribution mismatch.
\newblock In {\em Proceedings of the AAAI Conference on Artificial
  Intelligence.}, pages 3569--3576, 2020.

\bibitem{Doersch2015context}
C. {Doersch}, A. {Gupta}, and A.~A. {Efros}.
\newblock Unsupervised visual representation learning by context prediction.
\newblock In {\em Proceedings of the IEEE International Conference on Computer
  Vision}, pages 1422--1430, 2015.

\bibitem{dong2018tri}
WeiWang Dong-Dong~Chen and Zhi-Hua~Zhou WeiGao.
\newblock Tri-net for semi-supervised deep learning.
\newblock In {\em Proceedings of the International Joint Conferences on
  Artificial Intelligence}, pages 2014--2020, 2018.

\bibitem{Dosovitskiy2014discriminative}
Alexey Dosovitskiy, Jost~Tobias Springenberg, Martin Riedmiller, and Thomas
  Brox.
\newblock Discriminative unsupervised feature learning with convolutional
  neural networks.
\newblock In {\em Advances in Neural Information Processing Systems}, 2014.

\bibitem{gidaris2018unsupervised}
Spyros Gidaris, Praveer Singh, and Nikos Komodakis.
\newblock Unsupervised representation learning by predicting image rotations.
\newblock In {\em Proceedings of the International Conference on Learning
  Representations}, 2018.

\bibitem{Grandvalet2005Semi}
Yves Grandvalet and Yoshua Bengio.
\newblock Semi-supervised learning by entropy minimization.
\newblock In {\em Advances in Neural Information Processing Systems}, 2005.

\bibitem{guo2020safe}
Lan-Zhe Guo, Zhen-Yu Zhang, Yuan Jiang, Yu-Feng Li, and Zhi-Hua Zhou.
\newblock Safe deep semi-supervised learning for unseen-class unlabeled data.
\newblock In {\em International Conference on Machine Learning}, pages
  3897--3906. PMLR, 2020.

\bibitem{he2007graph}
Jingrui He, Jaime Carbonell, and Yan Liu.
\newblock Graph-based semi-supervised learning as a generative model.
\newblock In {\em Proceedings of the International Joint Conferences on
  Artificial Intelligence}, pages 2492--2497, 01 2007.

\bibitem{he2020momentum}
Kaiming He, Haoqi Fan, Yuxin Wu, Saining Xie, and Ross Girshick.
\newblock Momentum contrast for unsupervised visual representation learning.
\newblock In {\em Proceedings of the IEEE conference on Computer Vision and
  Pattern Recognition}, pages 9729--9738, 2020.

\bibitem{iscen2019label}
Ahmet Iscen, Giorgos Tolias, Yannis Avrithis, and Ondrej Chum.
\newblock Label propagation for deep semi-supervised learning.
\newblock In {\em Proceedings of the IEEE conference on Computer Vision and
  Pattern Recognition}, pages 5070--5079, 2019.

\bibitem{kall2007semi}
Lukas K{\"a}ll, Jesse~D Canterbury, Jason Weston, William~Stafford Noble, and
  Michael~J MacCoss.
\newblock Semi-supervised learning for peptide identification from shotgun
  proteomics datasets.
\newblock {\em Nature methods}, 4(11):923--925, 2007.

\bibitem{kashima2009link}
Hisashi Kashima, Tsuyoshi Kato, Yoshihiro Yamanishi, Masashi Sugiyama, and Koji
  Tsuda.
\newblock Link propagation: A fast semi-supervised learning algorithm for link
  prediction.
\newblock In {\em Proceedings of the SIAM International Conference on Data
  Mining}, pages 1100--1111. SIAM, 2009.

\bibitem{kipf2016semi}
Thomas~N. Kipf and Max Welling.
\newblock Semi-supervised classification with graph convolutional networks.
\newblock In {\em Proceedings of the International Conference on Learning
  Representations}, 2017.

\bibitem{kolesnikov2019revisiting}
Alexander Kolesnikov, Xiaohua Zhai, and Lucas Beyer.
\newblock Revisiting self-supervised visual representation learning.
\newblock In {\em Proceedings of the IEEE Conference on Computer Vision and
  Pattern Recognition}, June 2019.

\bibitem{krizhevsky2009learning}
A. Krizhevsky and G. Hinton.
\newblock Learning multiple layers of features from tiny images.
\newblock {\em Master's thesis, Department of Computer Science, University of
  Toronto}, 2009.

\bibitem{laine2016temporal}
Samuli Laine and Timo Aila.
\newblock Temporal ensembling for semi-supervised learning.
\newblock In {\em Proceedings of the International Conference on Learning
  Representations}, 2016.

\bibitem{le2015tiny}
Ya Le and Xuan Yang.
\newblock Tiny imagenet visual recognition challenge.
\newblock {\em CS 231N}, 7, 2015.

\bibitem{lee2018simple}
Kimin Lee, Kibok Lee, Honglak Lee, and Jinwoo Shin.
\newblock A simple unified framework for detecting out-of-distribution samples
  and adversarial attacks.
\newblock In {\em Advances in neural information processing systems},
  volume~31, 2018.

\bibitem{liang2017enhancing}
Shiyu Liang, Yixuan Li, and Rayadurgam Srikant.
\newblock Enhancing the reliability of out-of-distribution image detection in
  neural networks.
\newblock In {\em Proceedings of the International Conference on Learning
  Representations}, 2017.

\bibitem{luo2021consistency}
Huixiang Luo, Hao Cheng, Yuting Gao, Ke Li, Mengdan Zhang, Fanxu Meng, Xiaowei
  Guo, Feiyue Huang, and Xing Sun.
\newblock On the consistency training for open-set semi-supervised learning.
\newblock {\em arXiv preprint arXiv:2101.08237}, 2021.

\bibitem{maaten2008visualizing}
Laurens van~der Maaten and Geoffrey Hinton.
\newblock Visualizing data using t-sne.
\newblock {\em Journal of machine learning research}, 9(Nov):2579--2605, 2008.

\bibitem{Noroozi2016jiasaw}
Mehdi Noroozi and Paolo Favaro.
\newblock Unsupervised learning of visual representations by solving jigsaw
  puzzles.
\newblock In {\em Proceedings of the European Conference on Computer Vision},
  volume 9910, pages 69--84, 10 2016.

\bibitem{oliver2018realistic}
Avital Oliver, Augustus Odena, Colin Raffel, Ekin~D Cubuk, and Ian~J
  Goodfellow.
\newblock Realistic evaluation of deep semi-supervised learning algorithms.
\newblock In {\em Advances in Neural Information Processing Systems}, 2018.

\bibitem{Ostu1979A}
N Ostu, O Nobuyuki, and N Otsu.
\newblock A threshold selection method from gray- level histogram ieee
  transactions on systems.
\newblock {\em IEEE Trans.syst.man. \& Cybern}, 9(1):62--66, 1979.

\bibitem{qiao2018deep}
Siyuan Qiao, Wei Shen, Zhishuai Zhang, Bo Wang, and Alan Yuille.
\newblock Deep co-training for semi-supervised image recognition.
\newblock In {\em Proceedings of the European Conference on Computer Vision},
  pages 135--152, 2018.

\bibitem{Raghavan2007NearLT}
U. Raghavan, R. Albert, and S. Kumara.
\newblock Near linear time algorithm to detect community structures in
  large-scale networks.
\newblock {\em Physical review. E, Statistical, nonlinear, and soft matter
  physics}, 76 3 Pt 2:036106, 2007.

\bibitem{shrivastava2016training}
Abhinav Shrivastava, Abhinav Gupta, and Ross Girshick.
\newblock Training region-based object detectors with online hard example
  mining.
\newblock In {\em Proceedings of the IEEE conference on Computer Vision and
  Pattern Recognition}, pages 761--769, 2016.

\bibitem{Sohn2020FixMatch}
Kihyuk Sohn, David Berthelot, Chun~Liang Li, Zizhao Zhang, Nicholas Carlini,
  Ekin~D. Cubuk, Alex Kurakin, Han Zhang, and Colin Raffel.
\newblock Fixmatch: Simplifying semi-supervised learning with consistency and
  confidence.
\newblock In {\em Advances in Neural Information Processing Systems}, 2020.

\bibitem{2017Mean}
Antti Tarvainen and Harri Valpola.
\newblock Mean teachers are better role models: Weight-averaged consistency
  targets improve semi-supervised deep learning results.
\newblock In {\em Advances in Neural Information Processing Systems}, 2017.

\bibitem{triguero2015self}
Isaac Triguero, Salvador Garc{\'\i}a, and Francisco Herrera.
\newblock Self-labeled techniques for semi-supervised learning: taxonomy,
  software and empirical study.
\newblock {\em Knowledge and Information systems}, 42(2):245--284, 2015.

\bibitem{wang2013dynamic}
Bo Wang, Zhuowen Tu, and John~K Tsotsos.
\newblock Dynamic label propagation for semi-supervised multi-class multi-label
  classification.
\newblock In {\em Proceedings of the IEEE International Conference on Computer
  Vision}, pages 425--432, 2013.

\bibitem{xie2019unsupervised}
Qizhe Xie, Zihang Dai, Eduard Hovy, Thang Luong, and Quoc Le.
\newblock Unsupervised data augmentation for consistency training.
\newblock In H. Larochelle, M. Ranzato, R. Hadsell, M.~F. Balcan, and H. Lin,
  editors, {\em Advances in Neural Information Processing Systems}, volume~33,
  pages 6256--6268. Curran Associates, Inc., 2020.

\bibitem{yu2015lsun}
Fisher Yu, Ari Seff, Yinda Zhang, Shuran Song, Thomas Funkhouser, and Jianxiong
  Xiao.
\newblock Lsun: Construction of a large-scale image dataset using deep learning
  with humans in the loop.
\newblock {\em arXiv preprint arXiv:1506.03365}, 2015.

\bibitem{yu2020multi}
Qing Yu, Daiki Ikami, Go Irie, and Kiyoharu Aizawa.
\newblock Multi-task curriculum framework for open-set semi-supervised
  learning.
\newblock In {\em Proceedings of the European Conference on Computer Vision},
  2020.

\bibitem{zagoruyko2016wide}
Sergey Zagoruyko and Nikos Komodakis.
\newblock Wide residual networks.
\newblock {\em arXiv preprint arXiv:1605.07146}, 2016.

\bibitem{Zhai_2019_ICCV}
Xiaohua Zhai, Avital Oliver, Alexander Kolesnikov, and Lucas Beyer.
\newblock S4l: Self-supervised semi-supervised learning.
\newblock In {\em Proceedings of the IEEE International Conference on Computer
  Vision}, October 2019.

\bibitem{zhang01}
Hongyi Zhang, Moustapha Cisse, Yann~N Dauphin, and David Lopez-Paz.
\newblock mixup: Beyond empirical risk minimization.
\newblock In {\em Proceedings of the International Conference on Learning
  Representations}, 2017.

\bibitem{Zhu2002LearningFL}
Xiaojin Zhu and Zoubin Ghahramani.
\newblock Learning from labeled and unlabeled data with label propagation.
\newblock 2002.

\end{thebibliography}
	\end{document}